\documentclass[10pt,twocolumn,letterpaper]{article}

\usepackage{cvpr}              

\definecolor{cvprblue}{rgb}{0.21,0.49,0.74}
\usepackage[pagebackref,breaklinks,colorlinks,allcolors=cvprblue]{hyperref}
\usepackage{booktabs}
\usepackage{multirow}
\usepackage{amsmath}
\usepackage{amssymb}
\usepackage{placeins}
\usepackage{tabularx}
\usepackage[accsupp]{axessibility} 
\usepackage[table]{xcolor}

\title{V-Nutri: Dish-Level Nutrition Estimation from Egocentric Cooking Videos}

\author{
Chengkun Yue\\
Indiana University\\
{\tt\small yuech@iu.edu}
\and
Chuanzhi Xu\\
The University of Sydney\\
{\tt\small chuanzhi.xu@sydney.edu.au}
\and
Jiangpeng He\\
Indiana University\\
{\tt\small jhe2@iu.edu}
}

\begin{document}
\maketitle
\begin{abstract}
Nutrition estimation of meals from visual data is an important problem for dietary monitoring and computational health, but existing approaches largely rely on single images of the finally completed dish. This setting is fundamentally limited because many nutritionally relevant ingredients and transformations, such as oils, sauces, and mixed components, become visually ambiguous after cooking, making accurate calorie and macronutrient estimation difficult. In this paper, we investigate whether the cooking process information from egocentric cooking videos can contribute to dish-level nutrition estimation. First, we further manually annotated the HD-EPIC dataset and established the first benchmark for video-based nutrition estimation. Most importantly, we propose V-Nutri, a staged framework that combines Nutrition5K-pretrained visual backbones with a lightweight fusion module that aggregates features from the final dish frame and cooking process keyframes extracted from the egocentric videos. V-Nutri also includes a cooking keyframes selection module, a VideoMamba-based event-detection model that targets ingredient-addition moments. Experiments on the HD-EPIC dataset show that process cues can provide complementary nutritional evidence, improving nutrition estimation under controlled conditions. Our results further indicate that the benefit of process keyframes depends strongly on backbone representation capacity and event detection quality. Our code and annotated dataset will be released at \small{\url{https://github.com/K624-YCK/V-Nutri}}.

\end{abstract}

\section{Introduction}
\label{sec:intro}

\begin{figure}[t]
  \centering
  \includegraphics[width=\linewidth]{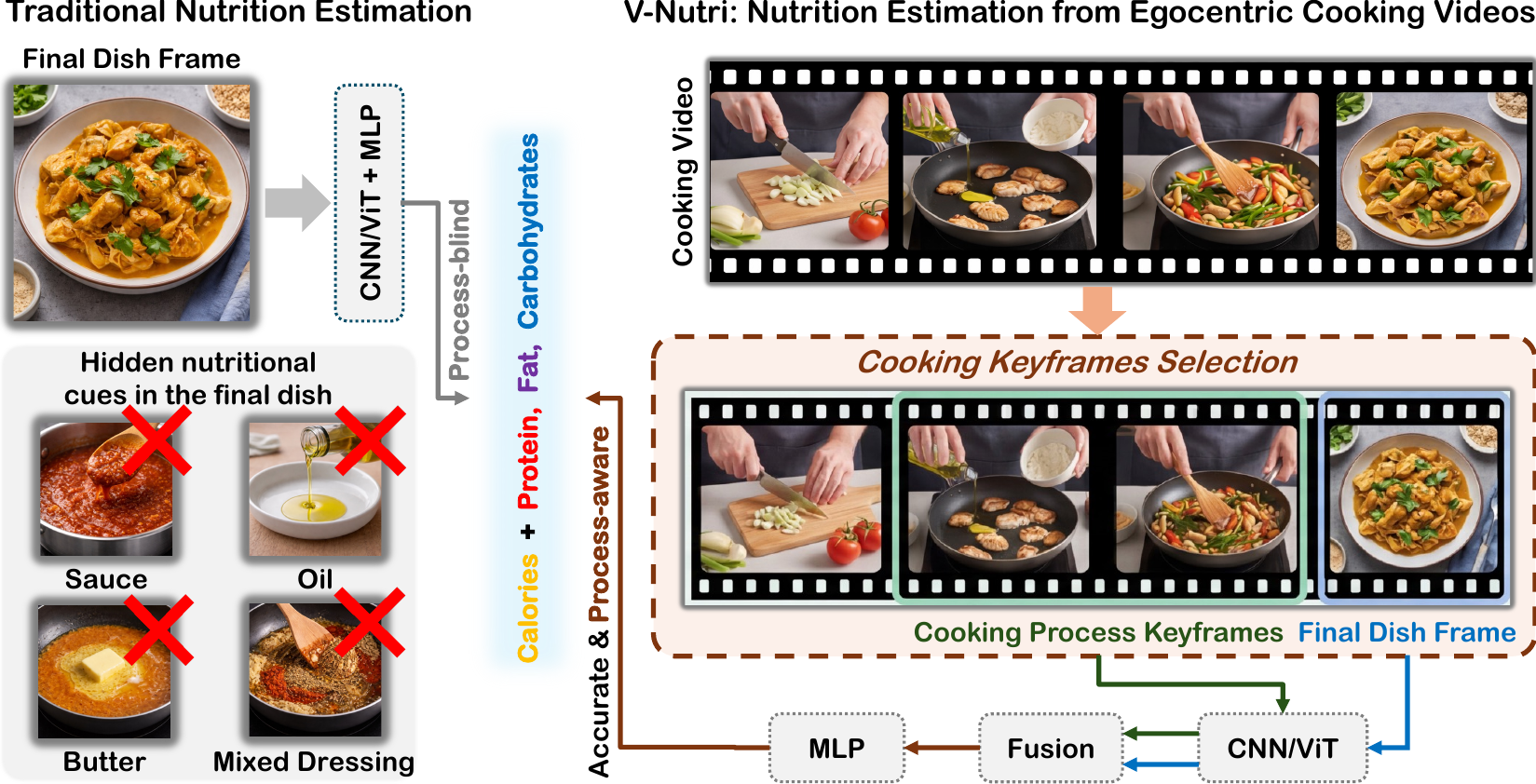}
  \caption{Traditional nutrition estimation is process-blind, relying only on the final dish image, where important ingredients may be hidden after cooking. \textbf{V-Nutri} combines process keyframes from egocentric cooking videos with the final dish frame for a process-aware estimation of calories and macronutrients.}
  \label{fig:teaser}
\end{figure}

Accurate estimation of food nutrition is essential for dietary monitoring, computational health, and personalized nutrition management. Ideally, a user could capture a meal with a camera, and an intelligent system would automatically estimate its calories, protein, fat, and carbohydrates \cite{myers2015im2calories}. Therefore, vision-based nutrition estimation has attracted growing attention \cite{zhang2025igsmnet,yan2025dietai24,vinod2026size, shao2023portion, vinod2024food, ma2024mfp3d, he2026physically, chen2026implicit} as an important problem in computational nutrition and food understanding.

Traditional nutritional assessment mainly relies on manual analysis, food diaries, or laboratory measurement \cite{ravelli2020traditional}. Although these approaches can provide relatively accurate nutritional information, they are labor-intensive, time-consuming, and difficult to scale to real-world daily use. These limitations have motivated increasing interest in automated visual nutrition estimation methods \cite{myers2015im2calories,thames2021nutrition5k,khlaisamniang2025decomposing}.

Recent advances in deep learning have significantly improved visual food analysis~\cite{mao2021visual, he2021online, he2025long}, and many existing methods estimate nutrition directly from images of the final dish \cite{he2020multi, shao2021towards, myers2015im2calories, thames2021nutrition5k, he2021end, ma2023improved, shao2022swinnutrition, khlaisamniang2025decomposing, zhang2025igsmnet}. While these methods have shown promising performance, they remain fundamentally constrained by the limited information available in a single post-cooking image. In realistic meals, many nutritionally important components, such as oils, sauces, dairy products, and mixed ingredients, are often absorbed, melted, or visually merged into the final dish. As a result, it is often difficult to accurately determine the calorie and macronutrient content merely by the final appearance.

We believe a promising direction is to move beyond static final-dish images and estimate nutrition from cooking videos, as shown in Figure \ref{fig:teaser}. Compared with a single post-cooking image, videos preserve substantially richer temporal nutritional evidence throughout food preparation, including ingredient identities, addition events, intermediate states, and procedural context that may disappear once the dish is completed. Moreover, this direction is becoming increasingly practical in real-world settings, as wearable cameras, action cameras, and other first-person recording devices are more frequently used for lifelogging, cooking documentation, and content creation \cite{rodin2021egocentricfuture,lo2024egodiet}. In particular, a recent research provides a suitable dataset, HD-EPIC \cite{perrett2025hdepic}, offering egocentric kitchen videos with recipe structure, temporal activity annotations (incomplete), and ingredient-level nutritional metadata. However, directly exploiting such videos is challenging. The recordings are long, unscripted, visually cluttered, and dominated by temporally redundant content, while nutritionally informative evidence is often sparse and concentrated in only a few key moments \cite{wu2019adaframe,lin2022ocsampler}.

In this research, we build a benchmark on the further annotated HD-EPIC dataset and propose a framework, \textbf{\underline{V}ideo-based \underline{Nutri}tion Estimation (\underline{V-Nutri})}, which includes a staged pipeline for dish-level process-aware nutrition estimation from egocentric cooking videos. We introduce a VideoMamba-based Process-frame Selection module pair (Cooking Keyframes Selection) \cite{li2024videomamba} to identify candidate nutritionally informative moments, such as ingredient-addition events. V-Nutri includes Nutrition5K-pretrained visual backbones \cite{thames2021nutrition5k} with a lightweight fusion module to combine evidence from selected process keyframes and the final dish image, predicting total calories, protein, fat, and carbohydrates. Our framework explicitly incorporates sparse process cues as complementary nutritional evidence, while avoiding the cost of dense long-video modeling as shown in many works \cite{pmlr-v139-bertasius21a,arnab2021vivit,liu2022video,li2022uniformerv2,tong2022videomae}.

Our contributions are summarized as follows:
\begin{itemize}
    \item We propose V-Nutri, the first process-aware framework for dish-level nutrition estimation from egocentric cooking videos, using process keyframes as complementary evidence to the final dish image.
    \item V-Nutri combines a VideoMamba-based Process-frame Selection module set (Cooking Keyframes Selection), Nutrition5K-pretrained visual backbones, and lightweight fusion and regression modules to predict calories, protein, fat, and carbohydrates. Experiments show that V-Nutri can improve the accuracy of nutrition estimation in many cases.
    \item We further annotate and structure the HD-EPIC dataset by complementing timestamp labels for cooking process keyframes and final dish keyframes, which can serve as the first video-based nutrition estimation benchmark.
\end{itemize}

\section{Related Work}
\label{sec:related}

\subsection{Nutrition Estimation from Static Food Images}

Vision-based nutrition estimation \cite{myers2015im2calories} predicts calories and nutrients from food images. Early work, such as Im2Calories \cite{myers2015im2calories} showed feasibility but highlighted a key challenge: inferring both food identity and quantity from limited visual cues \cite{myers2015im2calories}. Nutrition5K \cite{thames2021nutrition5k} introduced a stronger benchmark with multi-view observations and component-level weights, enabling direct nutrient regression \cite{thames2021nutrition5k}. Multimodal studies further reveal that final dish appearance often fails to capture full ingredient composition \cite{salvador2017recipe1m} and recipe structure \cite{salvador2019inverse}. To address this problem, prior work incorporates geometric cues \cite{lu2021dietaryassessment,ege2019realfoodsize} or ingredient semantics and external knowledge. For example, IGSMNet \cite{zhang2025igsmnet} and FLAVA \cite{feng2026flava} inject ingredient information, while DietAI24 \cite{yan2025dietai24} and ACETADA \cite{coburn2025acetada} leverage dietary databases and contextual metadata. Despite these advances, static-image methods remain limited to visible cues in the final dish \cite{myers2015im2calories,thames2021nutrition5k}. In many cooked meals, hidden ingredients and process-dependent transformations are not observable, motivating our exploration of process keyframes from cooking videos as complementary evidence.

\subsection{Long Video Understanding}

Long video understanding aims to model long temporal sequences efficiently without losing key context. Transformer-based models advanced this field by extending token-based representations to spatiotemporal data. TimeSformer \cite{pmlr-v139-bertasius21a} and ViViT \cite{arnab2021vivit} introduced factorized temporal modeling, while MViT \cite{fan2021multiscale}, Video Swin \cite{liu2022video}, UniFormer \cite{li2022uniformer}, and UniFormerV2 \cite{li2022uniformerv2} improved scalability via multiscale design and stronger priors.

Recent work emphasizes large-scale pretraining and general representations. VideoMAE \cite{tong2022videomae}, VideoMAE V2 \cite{wang2023videomaev2}, BEVT \cite{wang2022bevt}, and MaskFeat \cite{wei2022maskfeat} demonstrate effective masked pretraining, while VATT \cite{akbari2021vatt}, Frozen in Time \cite{bain2021frozen}, VideoCLIP \cite{xu2021videoclip}, OmniVL \cite{wang2022omnivl}, InternVideo \cite{wang2022internvideo}, and InternVideo2 \cite{wang2024internvideo2} extend to multimodal foundation models.

More recently, Mamba-based models \cite{gu2023mamba} replace quadratic attention with selective state-space modeling, enabling efficient long-range modeling. VideoMamba \cite{li2024videomamba} is particularly suitable for egocentric cooking videos \cite{perrett2025hdepic}, which are long and sparse in informative moments. In this work, we leverage such models for efficient process-frame selection rather than dense video modeling.

\begin{figure*}[t!]
    \centering
    \includegraphics[width=\textwidth]{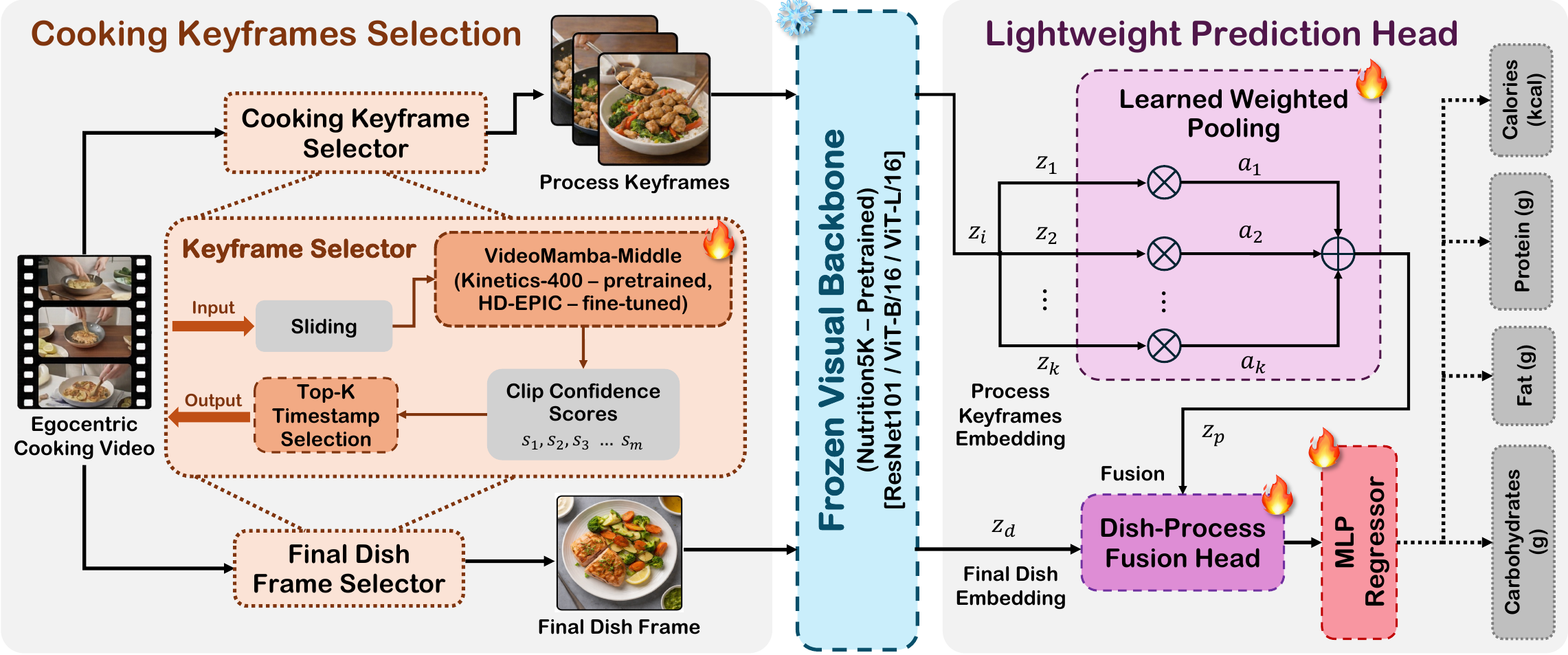}
    \caption{Overview of V-Nutri (Inference). Given an egocentric cooking video, the process keyframes and final dish frame are selected by the cooking keyframe selector and final dish frame selector. A frozen Nutrition5K-pretrained backbone (ResNet-101, ViT-B/16, or ViT-L/16) encodes each process keyframe into an embedding $z_1, \ldots, z_K$ and the final dish image into $z_d$. Learned attention weights $\alpha_1, \ldots, \alpha_K$ aggregate the process embeddings into a single representation $z_p$ through weighted pooling. The dish and process representations are fused and fed to an MLP regressor that predicts four nutritional values: calories, protein, fat, and carbohydrates.}
    \label{overview}
\end{figure*}

\subsection{Sparse Temporal Evidence Aggregation}

Many video tasks do not require dense processing of all frames. Prior work thus focuses on selecting informative subsets before heavy computation. ClipBERT \cite{lei2021clipbert} demonstrates effective sparse sampling for video-language learning, while OCSampler \cite{lin2022ocsampler}, NSNet \cite{xia2022nsnet}, and mmSampler \cite{hu2022mmsampler} compress long videos via salient evidence selection. AKS \cite{tang2025adaptive} and FFS \cite{buch2025flexible} further improve task-driven selection under limited budgets, emphasizing relevance and coverage. TokenLearner \cite{ryoo2021tokenlearner} and Video Token Merging \cite{lee2024videotokenmerging} show that learned token selection preserves useful information while reducing redundancy.

However, selection alone is insufficient. After selecting sparse evidence, models must aggregate it into a compact representation for prediction. In our work, we treat process keyframes as sparse cues of ingredients and transformations, aggregate them into a process representation, and fuse it with the final dish image for nutrition estimation. Unlike temporal action localization \cite{lin2019bmn} or long-video QA, our goal is not precise localization or full-sequence understanding, but efficient extraction and aggregation of process evidence. This aligns with our use of VideoMamba \cite{li2024videomamba} as a selection module to surface candidate frames rather than solve event localization directly.

\section{Methodology}
\label{sec:method}
\subsection{V-Nutri Framework Overview}

Figure~\ref{overview} shows the overall architecture of V-Nutri, which has a staged pipeline that yields a modular implementation with separate selection,  backbone, fusion, and regression components. Given an egocentric cooking video clip $c$, our goal is to predict the dish-level nutrition vectors $\mathbf{y}_c$ for Calories (kcal) and macronutrients: Protein (g), Fat (g), and Carbohydrates (g): 
\begin{equation}
\mathbf{y}_c = [y^{\text{kcal}}, y^{\text{protein}}, y^{\text{fat}}, y^{\text{carb}}] \in \mathbb{R}^4.
\end{equation}

After entering V-Nutri, the video clip first passes through our proposed cooking keyframe selector with the final dish frame selector, which selects the candidate cooking process keyframes and the final dish frame. Then, these process keyframes are fed into a frozen visual backbone pretrained on Nutrition5K~\cite{thames2021nutrition5k} to extract visual features. The extracted features are then aggregated through weighted pooling into a compact process representation, which is fused with the final dish embedding after being processed by the backbone. Finally, V-Nutri includes a lightweight regressor to predict dish-level nutrition.



Compared to the nutrition estimation methods that rely only on the final dish image, V-Nutri is designed to accurately estimate the dish-level nutrition with the in-process cooking frames. These frames may reveal critical events, such as ingredient additions and cooking transformations invisible in the completed dish.

\subsection{Cooking Keyframes Selection}





Although process keyframes can provide useful nutritional evidence, long egocentric cooking videos contain a large amount of irrelevant or redundant content. Dense processing over the full video is inefficient and may introduce noise. To address this issue, we propose the Cooking Keyframe Selector to detect candidate nutritionally informative events before downstream nutrition prediction.

Let the input cooking video be denoted as $v=\{f_t\}_{t=1}^{T}$, where $f_t$ is the frame at time index $t$ and $T$ is the total number of frames. Using a sliding-window strategy, we divide the video into a set of short temporal clips:
\begin{equation}
\mathcal{C}(v)=\{c_1,c_2,\dots,c_M\},
\end{equation}
where each clip $c_m$ spans 2 seconds and contains 16 frames at $224 \times 224$ resolution in our experimental settings.

The cooking keyframe selector is based on VideoMamba-Middle~\cite{li2024videomamba}, pretrained on Kinetics-400~\cite{carreira2017quo}, and fine-tuned on HD-EPIC \cite{perrett2025hdepic} to detect ingredient-addition events. For each clip $c_m$, the selector predicts a confidence score:
\begin{equation}
s_m = q_{\omega}(c_m), \qquad m=1,\dots,M,
\end{equation}
where $q_{\omega}$ denotes the selection model parameterized by $\omega$.

During inference, we apply the selector over sliding windows with a stride of 1 second and rank all clips according to their confidence scores. We then select the top-$K$ timestamps:
\begin{equation}
\mathcal{I}_{K} = \operatorname{TopK}\big(\{s_m\}_{m=1}^{M}\big).
\end{equation}
The corresponding candidate process keyframes are:
\begin{equation}
\mathcal{X}_{p}=\{x_i\}_{i=1}^{K},
\end{equation}
where each $x_i$ is the representative frame associated with a selected timestamp in $\mathcal{I}_{K}$.

In addition, we design a Final Dish Frame Selector using a similar formulation, except that it only outputs a single final-dish frame. Let $q_{\gamma}^{d}(c_m)$ denote the confidence that clip $c_m$ corresponds to the final-dish state. The selected final dish frame is obtained as
\begin{equation}
m^{*}=\arg\max_{m} q_{\gamma}^{d}(c_m),
\end{equation}
\begin{equation}
x_d = \phi(c_{m^{*}}),
\end{equation}
where $\phi(\cdot)$ extracts the representative frame from the selected clip.

\subsection{Frozen Visual Backbone}

A key challenge in our setting is the limited amount of usable nutritional supervision in HD-EPIC~\cite{perrett2025hdepic}. Directly training a large visual encoder on this dataset would make overfitting likely. To address this issue, we use Nutrition5K-pretrained visual backbones that were first fine-tuned on Nutrition5K~\cite{thames2021nutrition5k} and then kept frozen during downstream training on HD-EPIC \cite{perrett2025hdepic}.

Specifically, we compare three backbone architectures as visual encoders $f_\theta$: ResNet-101~\cite{he2016resnet}, ViT-B/16, and ViT-L/16~\cite{dosovitskiy2021vit}. These models produce embeddings of different dimensions (D): $D=2048$ for ResNet-101, $D=768$ for ViT-B/16, and $D=1024$ for ViT-L/16. Once the backbone is fixed, we extract a final-dish-image embedding and a process-keyframes embedding as:
\begin{equation}
\mathbf{z}_d = f_\theta(x_d), \qquad \mathbf{z}_i = f_\theta(x_i).
\end{equation}

We froze the visual backbone but only trained the lightweight prediction head on HD-EPIC, preserving the nutritional prior from Nutrition5K and enabling fair comparison across backbones while isolating the effect of process information.

\subsection{Lightweight Prediction Head: Process-Aware Aggregation and Fusion}

Visual evidence from only the final dish frame is often incomplete, while selected process keyframes can provide additional nutritional cues, but they are sparse and noisy. Therefore, we need to aggregate process evidence into a compact representation and combine it with the final-dish representation.

\subsubsection{Dish-Only Baseline}
\label{sec:dish-only}
First, we define a dish-only baseline. This model uses only the final-dish embedding and predicts nutrition through an MLP head $h_\phi$:
\begin{equation}
\hat{\mathbf{y}} = h_\phi(\mathbf{z}_d).
\end{equation}
This baseline provides a reference for subsequent comparisons and captures the information contained in the final dish image alone.

\subsubsection{Learned Weighted Pooling}

When process keyframes are available, we encode each frame with the same frozen backbone and then aggregate the resulting embeddings into a single process representation. Instead of treating all process keyframes equally, we learn attention weights that let the model focus on more informative moments. For each process embedding $\mathbf{z}_i$, we compute a scalar score:
\begin{equation}
s_i = g_\psi(\mathbf{z}_i),
\end{equation}
where $g_\psi$ is a two-layer attention network. We then normalize the scores with a softmax function:
\begin{equation}
\alpha_i = \frac{\exp(s_i)}{\sum_j \exp(s_j)}.
\end{equation}

Finally, we obtain the aggregated process representation $\mathbf{z}_p$ as a weighted sum:
\begin{equation}
\mathbf{z}_p = \sum_{i=1}^{N_c} \alpha_i \mathbf{z}_i.
\end{equation}
The model can assign larger weights to process keyframes that are more relevant to the final nutrition estimate. For comparison, we also consider a mean-pooling variant, where all keyframes receive equal weight, i.e., $\alpha_i = 1/N_c$.

\subsubsection{Dish-Process Fusion Head \& MLP Regressor}

After we obtain the dish representation $\mathbf{z}_d$ and the process representation $\mathbf{z}_p$, we need to combine them for the final prediction. We compare two fusion strategies.

\noindent{\textbf{Concatenation fusion.}}
The simplest strategy concatenates the two embeddings and inputs to an MLP regressor:
\begin{equation}
\hat{\mathbf{y}} = h_\phi([\mathbf{z}_d; \mathbf{z}_p]),
\end{equation}
where $[\cdot;\cdot]$ denotes vector concatenation. In this setting, $h_\phi$ is a three-layer MLP with dimensions $2D \rightarrow d_h \rightarrow d_h/2 \rightarrow 4$.

\noindent{\textbf{Gated projection fusion.}}
The second strategy first projects the dish and process embeddings into a shared hidden space and then mixes them with a learnable gate:
\begin{equation}
\hat{\mathbf{y}} = h_\phi\Big( \sigma(w) W_d \mathbf{z}_d + (1-\sigma(w)) W_e \mathbf{z}_p \Big),
\end{equation}
where $W_d, W_e \in \mathbb{R}^{d_h \times D}$ are learned projection matrices, $w$ is a scalar gate parameter initialized at $0.5$, and $\sigma(\cdot)$ is the sigmoid function. In this setting, $h_\phi$ is a two-layer MLP with dimensions $d_h \rightarrow d_h \rightarrow 4$.

Both fusion heads use ReLU activations and Dropout with $p=0.3$. We set $d_h=512$ in all experiments. These two strategies provide two alternative ways to combine dish and process representations.
\section{Video-based Nutri-Estimation Benchmark}
\label{sec:benchmark}

Existing nutrition estimation benchmarks such as Nutrition5K~\cite{thames2021nutrition5k} focus on static images of plated dishes. No prior work provides a benchmark for \emph{video-based} nutrition estimation that includes both temporal cooking process annotations and dish-level nutritional labels. In this section, we describe how we construct such a benchmark from HD-EPIC~\cite{perrett2025hdepic}, define the evaluation protocol, and explain our annotations for reproducibility.

\subsection{Source Dataset: HD-EPIC}

HD-EPIC~\cite{perrett2025hdepic} is a large-scale egocentric kitchen dataset designed for fine-grained activity understanding. It contains approximately 41 hours of cooking footage from 9 kitchens and 9 participants, recorded across 156 videos at 30\,fps with $1408{\times}1408$ fisheye resolution. The dataset provides hierarchical annotations, including recipe-level structure, ingredient metadata with nutritional values, and temporally grounded cooking actions (\eg, ``add,'' ``weigh,'' ``stir'').

Crucially, HD-EPIC was \emph{not} designed for nutrition estimation. The nutritional annotations are attached to individual ingredients rather than aggregated at the dish level, and no evaluation protocol exists for predicting dish-level nutrition from video. Our benchmark bridges this gap.

\subsection{Benchmark Construction}
\label{sec:benchmark_construction}

We extract recipe instances from HD-EPIC and construct a three-tier benchmark, summarized in \Cref{tab:benchmark_tiers}. The construction involves four steps:

\noindent{\textbf{Step 1: Recipe instance extraction.}
We identify 80 recipe instances spanning 69 unique recipes from the HD-EPIC annotations. Among them, 52 have complete ingredient-level nutrition labels for all ingredients used, The remaining 28 instances are missing nutritional values for one or more ingredients and cannot be used for regression. 22 of the 52 also have complete add-event timestamps for ground-truth evaluation. Each instance corresponds to one complete cooking session and includes: (a)~a final dish frame timestamp, (b)~a list of ingredients used, and (c)~when available, per-ingredient nutritional values and temporal ``add-event'' timestamps indicating when each ingredient was added during cooking.

\noindent{\textbf{Step 2: Nutritional target aggregation.}
For each recipe instance, we sum the per-ingredient nutritional values (calories, protein, fat, and carbohydrates) to obtain a dish-level target vector $\mathbf{y} \in \mathbb{R}^4$.

\noindent{\textbf{Step 3: Temporal annotation verification.}
 We use the 22 \emph{fully complete} instances for ground-truth evaluation, where the model receives the exact video frames at which ingredients were added. The other 30 instances have nutrient values but lack add-event timestamps for some ingredients, so their ground-truth process-frame sets are incomplete.

\noindent{\textbf{Step 4: Manual annotation refinement.}
The original HD-EPIC timestamps are designed for activity recognition and do not optimize for visual clarity of ingredients. We therefore manually re-annotate all 52 regression-ready instances. For each add-event, The annotator watches the corresponding video segment and selects the single frame where: (1)~the ingredient is most clearly visible, (2)~lighting conditions are best, and (3)~the ``adding'' action is evident. For each final dish frame, the annotator selects the frame where the completed dish is most clearly and completely visible. This process updated 372 add-event timestamps and 52 dish-frame timestamps across the benchmark.

\begin{table}[t]
  \centering
  \caption{Three-tier structure of the HD-EPIC nutrition estimation benchmark. Each tier is a subset of the one above.}
  \label{tab:benchmark_tiers}
  \small
  \begin{tabularx}{\columnwidth}{lcX}
    \toprule
    Tier & Count & Definition \\
    \midrule
    All instances     & 80 & Recipe instances extracted from HD-EPIC (69 unique recipes) \\
    \addlinespace
    Regression-ready  & 52 & All ingredients have nutritional labels $\to$ dish-level target available \\
    \addlinespace
    Fully complete    & 22 & All ingredients have both nutritional labels \emph{and} add-event timestamps \\
    \bottomrule
  \end{tabularx}
\end{table}

\subsection{Data Distribution}
\label{sec:data_distribution}

\Cref{fig:nutrition_dist} shows the distribution of the four nutrient values across the 52 regression-ready instances. All four nutrients exhibit right-skewed distributions with high variance. Calories range from 0 to 4{,}792\,kcal, protein from 0 to 175\,g, fat from 0 to 329\,g, and carbohydrates from 0 to 616\,g.

Four instances have zero values for all nutrients (corresponding to water or tea preparations). Several additional instances have near-zero values for individual macronutrients (\eg, protein $<$\,5\,g for dishes consisting primarily of oils or fats). This distribution implicates that the presence of zero and near-zero targets makes mean absolute percentage error (MAPE) unreliable, as even small absolute errors produce extreme percentage errors when the denominator approaches zero. We therefore adopt per-nutrient MAE as the primary evaluation metric.

\begin{figure}[t]
    \centering
    \includegraphics[width=\columnwidth]{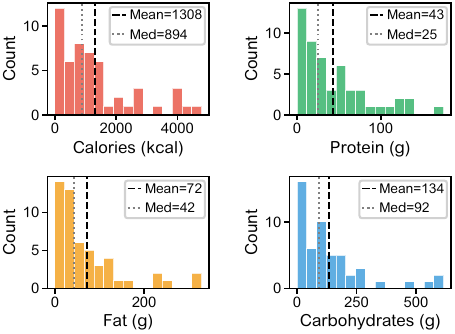}
    \caption{Nutrition distribution of 52 benchmark instances. Violin plots show density with individual points; dashed and dotted lines denote the mean and median. All nutrients are right-skewed, and four instances have zero values across all nutrients.}
    \label{fig:nutrition_dist}
\end{figure}

\subsection{Evaluation Protocol}
\label{sec:eval_protocol}

\noindent{\textbf{Cross-validation.}
We use stratified 5-fold cross-validation at the recipe level (seed = 42), ensuring each recipe appears in exactly one test fold. Each fold contains 40–43 training and 9–12 test instances with nutrient values. Targets are z-score normalized within each fold using training-set statistics to prevent data leakage and balance the contribution of the four nutrients during training.

\noindent{\textbf{Metrics.}
We report per-nutrient MAE for calories (kcal), protein (g), fat (g), and carbohydrates (g). Since these nutrients differ greatly in scale, we report them separately rather than combining them into a single aggregate score.

\noindent{\textbf{Process-frame sampling strategies.}
To evaluate how process information affects nutrition estimation, we define six sampling strategies:
\begin{itemize}
    \item \emph{Ground Truth (GT)}: uses all manually annotated add-event timestamps (available for the 22 fully complete instances; partial for the remaining 30).
    \item \emph{Predicted top-$K$ (Pred-K)}: selects the $K$ highest-confidence frames with the cooking keyframes selection.
    \item \emph{Predicted-all (Pred-all)}: select all frames above the confidence threshold with the cooking keyframes selection.
    \item \emph{Random-$K$ (Rand-K)}: draws $K$ frames uniformly at random from the video duration.
    \item \emph{Uniform-$K$ (Uni-K)}: places $K$ frames at equal temporal intervals across the video.
    \item \emph{Dish-only}: uses only the final dish frame (no process keyframes), serving as the baseline.
\end{itemize}
For predicted, random, and uniform strategies, we compare frame budgets $K \in \{20, 50, 200\}$.
\section{Experiments}
\label{sec:experiments}

\subsection{Datasets, Metrics, and Implementation}

\noindent{\textbf{Nutrition5K.}}
We use Nutrition5K~\cite{thames2021nutrition5k} to fine-tune image backbones and compare their performance before downstream evaluation on HD-EPIC.
The dataset contains about 5k real dishes with nutritional annotations and visual streams, including overhead RGB-D images and side-angle videos. In our experiments, we use overhead RGB images and middle frames extracted from side-angle videos from four cameras per dish. This setting gives 17{,}883 training samples and 3{,}108 validation samples, covering 4{,}059 training dishes and 709 validation dishes.

\noindent{\textbf{HD-EPIC.}}
We evaluate our method on the HD-EPIC nutrition estimation benchmark described in \Cref{sec:benchmark}.  Dataset statistics, tier definitions, and sampling strategies are detailed in \Cref{sec:benchmark_construction,sec:data_distribution,sec:eval_protocol}. 


\noindent{\textbf{Implementation Details.}}
All models are implemented in PyTorch.
For the HD-EPIC pipeline, each backbone is loaded from its Nutrition5K-fine-tuned checkpoint and used as a frozen feature extractor.
The embedding dimension is $D{=}2048$ for ResNet101, $D{=}768$ for ViT-B/16, and $D{=}1024$ for ViT-L/16.
We pre-extract embeddings at 1\,fps for all 80 recipe videos and store them on disk.
This design allows fast evaluation across all methods and folds.

We evaluate two fusion architectures, concatenation and gated projection, as described in \cref{sec:method}, with $d_h{=}512$.
The Ground Truth validation uses gated fusion.
The main 5-fold evaluation uses gated projection by default, and we compare both fusion architectures in an ablation study.
We z-score normalize training targets within each fold and denormalize predictions before computing MAE.
We train fusion heads with Smooth~L1 loss and Adam optimizer ($\text{lr}{=}10^{-3}$) for 50 epochs per fold.

For the cooking keyframes selection, we fine-tune VideoMamba-Middle~\cite{li2024videomamba}, pretrained on Kinetics-400, on the ingredient-addition detection task.
We use AdamW with learning rate $10^{-4}$, weight decay 0.05, and 3 epochs of linear warm-up from $10^{-6}$.
Training runs for 30 epochs with batch size 16.
Positive clips receive class weight 2.0.
Each clip is 2.0\,s long and contains 16 frames at $224{\times}224$, with a 2:1 negative-to-positive sampling ratio.
We store data in LMDB for efficient loading.
The best checkpoint achieves 62.7\% F1 on the fold-0 test split at epoch 14.
During inference, we apply a sliding window with 1-second stride to generate per-clip selection scores across the full test video.

All training and inference experiments are conducted on 2× NVIDIA H100 GPUs.


\subsection{Backbone Comparison on Nutrition5K}
\label{sec:backbone_selection}

We first compare three backbone architectures, namely ResNet101, ViT-B/16, and ViT-L/16, fine-tuned on Nutrition5K using overhead and side-angle images, with MSE loss on per-nutrient z-score normalized targets, data augmentation (RandomResizedCrop, HorizontalFlip, and ColorJitter), Dropout($p{=}0.3$) on the final layer, AdamW ($\text{lr}{=}10^{-4}$, weight decay $10^{-4}$), linear warmup for 5 epochs, cosine annealing, and early stopping with patience 30.

\Cref{tab:backbone} reports the Nutrition5K validation results.
ViT-L/16 achieves the lowest validation loss (2.203), while ResNet101 gives the best calorie MAE (73.2\,kcal).
ResNet101 also gives the best fat, carbohydrate, and protein MAE on this validation split. This may suggest that for datasets like Nutrition5K, which rely on final dish images, simpler models may yield better results. However, all three backbone architectures are capable of effectively estimating nutritional values.

\begin{table}[t]
  \centering
  \caption{Backbone comparison on Nutrition5K 
    (training with z-score normalization and augmentation). The best result per column is highlighted in \textbf{bold}. (Cal in kcal; Prot/Fat/Carb in g)}
  \label{tab:backbone}
  \small
  \resizebox{\linewidth}{!}{
  \begin{tabular}{lcccccc}
    \toprule
    Model & Epoch & Val Loss & Cal & Fat & Carb & Prot \\
    \midrule
    ResNet101  & 48 & 2.308 & \textbf{73.2}  & \textbf{5.60} & \textbf{6.63} & \textbf{5.14} \\
    ViT-B/16   & 4  & 2.559 & 88.0  & 7.00 & 7.95 & 6.52 \\
    ViT-L/16   & 31 & \textbf{2.203} & 77.7  & 5.76 & 6.81 & 5.19 \\
    \bottomrule
  \end{tabular}
  }
\end{table}

\subsection{Validation of Process Keyframes Utility}
\label{sec:Ground Truth}

We conduct a validation on HD-EPIC using annotated ground-truth final dish frame and process keyframes (without Cooking Keyframes Selection) to test whether ingredient-addition events provide cues beyond the final dish. Using the same pretrained backbones and normalization, we freeze features and train a simple fusion head. We conduct comparative experiments using only the final dish frame, and using process keyframes with mean pooling and weighted pooling.

\Cref{tab:Ground Truth} reports per-nutrient MAE as 5-fold mean for all three backbones.
All three backbones show improved calorie estimation when adding process keyframes.
For ResNet101, mean pooling reduces calories MAE from 698 to 572\,kcal.
For ViT-B/16, mean pooling reduces calories MAE from 672 to 569\,kcal.
For ViT-L/16, weighted pooling reduces calories MAE from 711 to 551\,kcal.

All macronutrients were improved across all backbones.
This validation supports our main hypothesis: process keyframes in videos can contain nutritional evidence invisible in the final dish image, and can contribute to the dish-level nutrition estimation.

\begin{table}[t]
  \centering
  \caption{Validation of Process Keyframes Utility with ground-truth process keyframes. With final dish frame is marked as D; with process keyframes is marked as P. The best result per column is highlighted in \textbf{bold}. (Cal in kcal; Prot/Fat/Carb in g)}
  \label{tab:Ground Truth}
  \scriptsize
  \setlength{\tabcolsep}{2.5pt}
  \resizebox{\linewidth}{!}{
  \begin{tabular}{l cccc cccc cccc}
    \toprule
    & \multicolumn{4}{c}{\textbf{ResNet101}}
    & \multicolumn{4}{c}{\textbf{ViT-B/16}}
    & \multicolumn{4}{c}{\textbf{ViT-L/16}} \\
    \cmidrule(lr){2-5} \cmidrule(lr){6-9} \cmidrule(lr){10-13}
    Method & Cal & Pr & Fat & Cb & Cal & Pr & Fat & Cb & Cal & Pr & Fat & Cb \\
    \midrule
    D(GT)-only        & 698 & 27.8 & 37.7 & 101
                     & 672 & 26.5 & 39.5 & 88.5
                     & 711 & 32.5 & 45.9 & 89.9 \\
    D+P\ (mean) & \textbf{572} & 24.6 & \textbf{32.9} & 84.3
                     & \textbf{569} & \textbf{22.2} & 33.7 & \textbf{80.3}
                     & 596 & 28.3 & 43.0 & 81.8 \\
    D+P\ (wtd.) & 582 & \textbf{22.3} & 35.9 & \textbf{73.9}
                     & 571 & 23.1 & \textbf{30.9} & 91.1
                     & \textbf{551} & \textbf{22.5} & \textbf{39.4} & \textbf{74.4} \\
    \bottomrule
  \end{tabular}
  }
\end{table}

\begin{table}[t]
  \centering
  \caption{5-fold cross-validation results (mean MAE) for ablation study on sampling strategies. The best result per column is highlighted in \textbf{bold}. (Cal in kcal; Prot/Fat/Carb in g)}
  \label{tab:main_ab_updated}
  \scriptsize
  \setlength{\tabcolsep}{2.5pt}
  \resizebox{\linewidth}{!}{%
  \begin{tabular}{l cccc cccc cccc}
    \toprule
    & \multicolumn{4}{c}{\textbf{ResNet101}} 
    & \multicolumn{4}{c}{\textbf{ViT-B/16}} 
    & \multicolumn{4}{c}{\textbf{ViT-L/16}} \\
    \cmidrule(lr){2-5} \cmidrule(lr){6-9} \cmidrule(lr){10-13}
    Method & Cal & Pr & Fat & Cb & Cal & Pr & Fat & Cb & Cal & Pr & Fat & Cb \\
    \midrule
    \rowcolor{gray!15} GT & 582 & 22.3 & 35.9 & 73.9
                     & 571 & 23.1 & 30.9 & 91.1
                     & 551 & 22.5 & 39.4 & 74.4 \\
    \midrule
    Dish-only      & 797 & 29.8 & 46.1 & 92.2 & 666 & 26.3 & 37.4 & 94.8 & 729 & 31.1 & \textbf{43.4} & 86.1 \\
    \midrule
    Pred-20        & \textbf{763} & \textbf{28.3} & 44.3 & \textbf{86.7} & 659 & \textbf{22.8} & \textbf{35.7} & 91.4 & 716 & 29.4 & 44.5 & 76.6 \\
    Pred-50        & 765 & 28.5 & \textbf{44.2} & 86.9 & 663 & 24.1 & 36.8 & 88.9 & 693 & 29.0 & 43.5 & 74.7 \\
    Pred-200       & 767 & 28.6 & 44.3 & 86.9 & 657 & 24.8 & 36.7 & \textbf{87.5} & \textbf{682} & \textbf{28.9} & 44.2 & \textbf{73.1} \\
    Pred-all       & 768 & 28.6 & 44.4 & 87.0 & \textbf{642} & 24.4 & 36.0 & 88.0 & 690 & 29.1 & 44.8 & 73.6 \\
    \midrule
    Rand-20        & 803 & 30.8 & 46.9 & 94.1 & 731 & 28.4 & 41.6 & 97.6 & 761 & 33.0 & 47.9 & 82.3 \\
    Rand-50        & 799 & 30.5 & 46.6 & 92.9 & 716 & 27.7 & 40.4 & 95.1 & 754 & 31.9 & 47.5 & 80.3 \\
    Rand-200       & 794 & 30.2 & 46.3 & 91.9 & 704 & 26.9 & 39.2 & 92.8 & 746 & 31.6 & 47.2 & 79.7 \\
    \midrule
    Uni-20         & 806 & 31.1 & 47.3 & 94.6 & 725 & 28.1 & 41.2 & 96.5 & 767 & 32.6 & 46.7 & 84.1 \\
    Uni-50         & 796 & 30.3 & 46.7 & 93.4 & 709 & 27.5 & 39.8 & 94.1 & 757 & 32.2 & 47.3 & 82.1 \\
    Uni-200        & 791 & 29.9 & 46.0 & 91.7 & 698 & 26.5 & 38.8 & 92.5 & 751 & 31.8 & 47.8 & 80.8 \\
    \bottomrule
  \end{tabular}
  }
\end{table}

\subsection{Ablation Study on Sampling Strategies}
\label{sec:main_results}

Table~\ref{tab:main_ab_updated} reports the 5-fold cross-validation ablation results of all sampling strategies with weighted pooling and gated fusion across the three backbones. Overall, the utility of process keyframes is strongly backbone-dependent, and the proposed prediction-based sampling strategies relying on our proposed Cooking Keyframes Selection module can consistently outperform random and uniform baselines. Ground-Truth (GT) keyframes serve as an upper-bound reference and generally provide the strongest overall performance, while the gap between GT and prediction-based sampling reflects the remaining room for improvement in process-frame selection. The Dish-only baseline (\Cref{sec:dish-only}), which uses only the predicted final dish frame as input, performs worse than our proposed architecture that includes predicting process keyframes, but slightly better than randomly or uniformly selecting process keyframes.

\begin{figure}[t]
  \centering
  \includegraphics[width=\columnwidth]{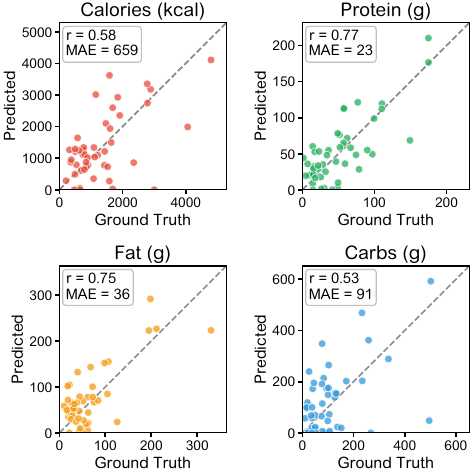}
  \caption{Predicted vs.\ ground-truth nutrient values for the ViT-B/16 with Pred-20 strategy. Dashed line: perfect prediction. $r$: Pearson correlation.}
  \label{fig:scatter}
\end{figure}

On ResNet101, among sampling strategies, prediction-based sampling consistently outperforms dish-only, random, and uniform baselines. Pred-20 gives the best calorie, protein, and carbohydrate MAE (763\,kcal, 28.3\,g, and 86.7\,g), while Pred-50 yields the best fat MAE (44.2\,g). On ViT-B/16, the prediction-based samplings lead as well. Pred-all achieves the best calorie MAE (642\,kcal), Pred-20 performs best on protein and fat (22.8\,g and 35.7\,g), and Pred-200 gives the best carbohydrate MAE (87.5\,g). On ViT-L/16, prediction-based sampling again improves over dish-only for calories, protein, and carbohydrates, while the trend for fat is less consistent. Pred-200 achieves the best automatic results for calories, protein, and carbohydrates (682\,kcal, 28.9\,g, and 73.1\,g), and Dish-only gives the best fat MAE (43.4\,g). In particular, Pred-200 reduces carbohydrate MAE by 13.0\,g relative to dish-only.

We further observe that increasing the number of sampled frames does not consistently improve performance. On ResNet101 and ViT-B/16, Pred-20 already achieves the best results on most metrics, while adding more frames brings limited gains. By contrast, ViT-L/16 benefits more from larger budgets of sampling frames for higher calories, protein, and carbohydrates. Random and uniform samplings improve only marginally as more frames are used, but remain consistently worse than prediction-based sampling. These results suggest that performance depends more on selecting informative process keyframes than on simply increasing the number of sampled frames.

\Cref{fig:scatter} shows the per-instance predictions versus ground truth for the Pred-20 strategy on ViT-B/16, a setting that performs relatively well across all four nutrients. The Pearson correlation results generally show positive trends and are shown better on Protein and Fat. However, there is still a clear gap from ideal prediction, reflecting the challenge of learning from only 52 training samples with a highly skewed distribution.

\begin{table}[!t]
  \centering
  \caption{Fusion architecture ablation (Calories MAE, 5-fold mean). The best result per column is highlighted in \textbf{bold}.}
  \label{tab:fusion_ablation}
  \small
  \setlength{\tabcolsep}{4pt}
  \begin{tabular}{llccc}
    \toprule
    Backbone & Fusion & Dish-only & Ground Truth & Pred top-20 \\
    \midrule
    \multirow{2}{*}{ResNet101}
      & Concat & \textbf{785} & \textbf{568} & 768 \\
      & Gated  & 797 & 582 & \textbf{763} \\
    \midrule
    \multirow{2}{*}{ViT-B/16}
      & Concat & 674 & \textbf{565} & 674 \\
      & Gated  & \textbf{666} & 571 & \textbf{659} \\
    \midrule
    \multirow{2}{*}{ViT-L/16}
      & Concat & \textbf{707} & 573 & \textbf{682} \\
      & Gated  & 729 & \textbf{551} & 716 \\
    \bottomrule
  \end{tabular}
\end{table}

\subsection{Effect of Fusion Architecture}

We compare concatenation and gated projection fusion in \Cref{tab:fusion_ablation}. Fusion performance depends on both backbone and sampling strategy.
On ResNet101, concatenation is better for Dish-only and Ground Truth, while gated projection slightly improves predicted top-20.
On ViT-B/16, gated projection performs best for Dish-only and predicted top-20, whereas concatenation is slightly better under Ground Truth.
On ViT-L/16, concatenation outperforms for Dish-only and predicted top-20, while gated projection achieves the best Ground Truth result.

Overall, no fusion strategy is consistently optimal.
Concatenation is more reliable when process frames are well-aligned (Ground Truth), while gated projection can better handle noisier inputs in some cases.
These results indicate that the optimal fusion design depends jointly on backbone capacity and process-frame quality.
\FloatBarrier
\section{Conclusion}
\label{sec:conclusion}


We present V-Nutri, a process-aware framework for dish-level nutrition estimation from egocentric cooking videos. By combining sparse process keyframes with the final dish image, it addresses the limitations of static-image-based methods. We also establish the first benchmark for video-based nutrition estimation by further annotating HD-EPIC, enabling reproducible evaluation. Experiments show that process cues provide complementary nutritional information, especially for macronutrients, though their effectiveness depends on backbone representation quality and event selection accuracy. Limitations remain, including imperfect event detection and limited dataset scale. Future work will improve selection models, scale data, and explore stronger representations. Overall, this work highlights the importance of temporal process modeling and offers a promising direction for video-based nutrition estimation.
{
    \small
    \bibliographystyle{ieeenat_fullname}
    \bibliography{main}
}

\end{document}